\documentclass[11pt]{article}

\usepackage[final]{acl}

\usepackage{times}
\usepackage{latexsym}

\usepackage[T1]{fontenc}

\usepackage[utf8]{inputenc}

\usepackage{microtype}

\usepackage{inconsolata}

\usepackage{graphicx}
\usepackage{tcolorbox}
\usepackage{colortbl}
\usepackage{geometry}
\usepackage{amsmath}
\usepackage{algorithm}
\usepackage{algorithmic}
\usepackage{multirow}
\usepackage{amssymb}
\usepackage{booktabs}
\usepackage{array}
\usepackage{cellspace}
\usepackage{svg}
\title{FreezeEmpath: Efficient Training for Empathetic Spoken Chatbots with Frozen LLMs}

\author{
    Yun Hong\textsuperscript{\rm 1,2,3},
    Yan Zhou\textsuperscript{\rm 1,2,3},
    Yang Feng\textsuperscript{\rm 1,2,3}\footnotemark[2] \\
    \textsuperscript{\rm 1}{Key Laboratory of Intelligent Information Processing, Institute of Computing Technology,} \\ Chinese Academy of Sciences (ICT/CAS) \textsuperscript{\rm 2} {State Key Laboratory of AI Safety,} \\ Institute of Computing Technology, Chinese Academy of Sciences \\
    \textsuperscript{\rm 3} {University of Chinese Academy of Sciences, Beijing, China} \\
    \texttt{\href{mailto:hongyun25@mails.ucas.ac.cn}{hongyun25@mails.ucas.ac.cn},  \href{mailto:fengyang@ict.ac.cn}{fengyang@ict.ac.cn}}
}

\begin{document}
\maketitle
\renewcommand{\thefootnote}{\fnsymbol{footnote}} 
\footnotetext[2]{Corresponding author: Yang Feng.} 
\renewcommand{\thefootnote}{\arabic{footnote}}
\begin{abstract}
Empathy is essential for fostering natural interactions in spoken dialogue systems, as it enables machines to recognize the emotional tone of human speech and deliver empathetic responses. Recent research has made significant progress in developing empathetic spoken chatbots based on large language models (LLMs). However, several challenges still exist when training such models, including reliance on costly empathetic speech instruction data and a lack of emotional expressiveness in the generated speech. Finetuning LLM with cross-modal empathetic instruction data may also lead to catastrophic forgetting and a degradation of its general capability. To address these challenges, we propose \textbf{FreezeEmpath}, an end-to-end empathetic spoken chatbot trained in a simple and efficient manner. The entire training process relies solely on existing speech instruction data and speech emotion recognition (SER) data, while keeping the LLM's parameters frozen. Experiments demonstrate that FreezeEmpath is able to generate emotionally expressive speech and outperforms other empathetic models in empathetic dialogue, SER, and SpokenQA tasks, demonstrating the effectiveness of our training strategy. \footnote{\url{https://github.com/ictnlp/FreezeEmpath}}
\end{abstract}

\section{Introduction}
Empathy plays a crucial role in human–machine spoken interaction, allowing machines to capture emotional cues embedded in speech prosody, understand users’ underlying psychological states, and produce responses that are both contextually appropriate and emotionally empathetic. As LLMs have demonstrated powerful human-computer interaction capabilities in recent years \cite{gpt4o, yang2025qwen3technicalreport}, prior studies have incorporated the speech modality into LLMs, leading to the development of spoken chatbots that support natural spoken interactions with users \cite{zeng2024glm4voiceintelligenthumanlikeendtoend, fang-etal-2025-llama}. To further enhance the empathetic capacity of dialogue systems, recent research has explored empathetic spoken chatbots that can not only understand users’ speech but also generate emotionally appropriate and empathetic speech responses \cite{wang2025opens2sadvancingfullyopensource, geng2025osumechatenhancingendtoendempathetic}.

Despite promising results, several challenges remain in training such empathetic spoken chatbots. The most significant challenge lies in the scarcity of empathetic speech instruction data. To construct such data, existing methods typically leverage a powerful LLM \cite{deepseekai2025deepseekv3technicalreport, yang2025qwen3technicalreport} to generate empathetic textual dialogues and corresponding emotion labels in empathetic scenarios. These textual dialogues are then converted into empathetic speech instruction data using emotion-controllable text-to-speech (TTS) models \cite{du2024cosyvoice2scalablestreaming}. In addition to the complexity and high cost of the data construction process, the textual dialogues generated in this way lack sufficient content diversity, leading to poor generalization capability of the trained model. Furthermore, finetuning LLM on such cross-modal data could also potentially cause catastrophic forgetting, resulting in a degradation of the model’s general ability. Another challenge is that existing empathetic spoken chatbots often generate speech with insufficient emotional expressiveness, failing to effectively convey the intended emotional cues.

\begin{figure}[t]
  \includegraphics[width=1.00\linewidth]{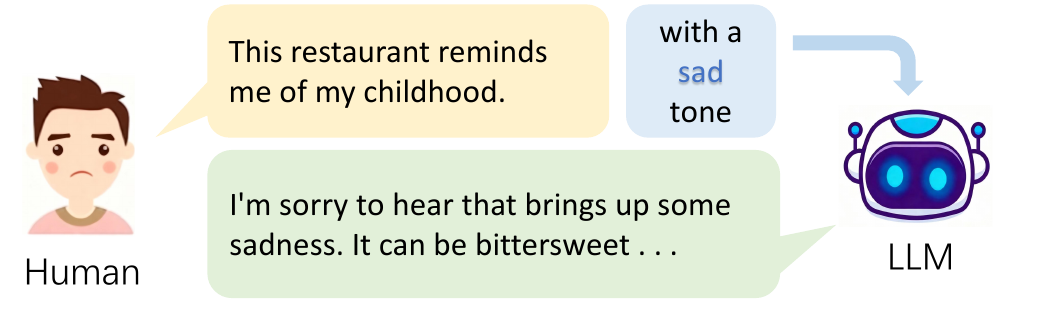}
  \caption {Demonstration of the LLM’s inherent empathetic capability.}
  \label{intro}
\end{figure}

To overcome the above challenges, we propose \textbf{FreezeEmpath}, an empathetic spoken chatbot trained in an efficient manner, with the base LLM being frozen. The key insight of our method is that LLMs already possess inherent empathetic capability. If the frozen LLM is explicitly provided with the emotional tone of the speech, it can naturally generate a high-quality empathetic response, as shown in Figure \ref{intro}. 

To leverage the LLM’s inherent empathetic capability, we adopt a semantic–emotion decoupled encoding strategy that separately encodes features related to the semantic content and the emotional tone of speech. This is accomplished by using a semantic speech adapter and an emotion extractor to obtain semantic and emotional features from a shared speech encoder. We then design two training stages—\textbf{semantic alignment} and \textbf{emotional alignment}—to align these features with the LLM’s embedding space, thereby transferring the LLM’s inherent empathetic ability from the text modality to the speech modality. Unlike prior approaches, our decoupled encoding strategy eliminates the dependence on real empathetic spoken dialogue data. By leveraging existing speech instruction data and SER data, we can construct pseudo-empathetic speech instruction data through a self-instruct process that exploits the LLM’s inherent empathetic capability. Compared to manually collecting or synthesizing real speech data, our approach is significantly more cost-effective and highly scalable. To further enable the model to generate emotionally expressive speech, we adopt a \textbf{speech generation} training stage, introducing a streaming speech decoder that generates speech tokens based on the hidden states of the LLM. By introducing effective emotional supervision, these speech tokens contain both the semantic content of the model's response and the emotional prosody information that aligns with the semantics. These tokens are then converted into emotional speech using a token-to-wav module. The base LLM remains frozen in the whole training process, ensuring that its general capabilities are not compromised.

Experimental results demonstrate that our model achieves remarkable performance across a wide range of tasks, including speech emotion recognition, spoken question answering, speech instruction following, and empathetic conversation, validating the effectiveness and efficiency of our approach.

Our main contributions to training an empathetic spoken chatbot can be summarized as follows:
\begin{itemize}
    \item We use a semantic–emotion decoupled encoding strategy and a two-stage alignment training method to transfer the LLM's inherent empathetic capability to speech modality.
    \item Our entire training process relies only on existing speech instruction data and SER data, with no need for carefully curated empathetic speech instruction data.
    \item The LLM remains completely frozen throughout the entire training process, preserving its knowledge and general capability.
\end{itemize}

\begin{figure*}[t]
  \includegraphics[width=0.95\linewidth]{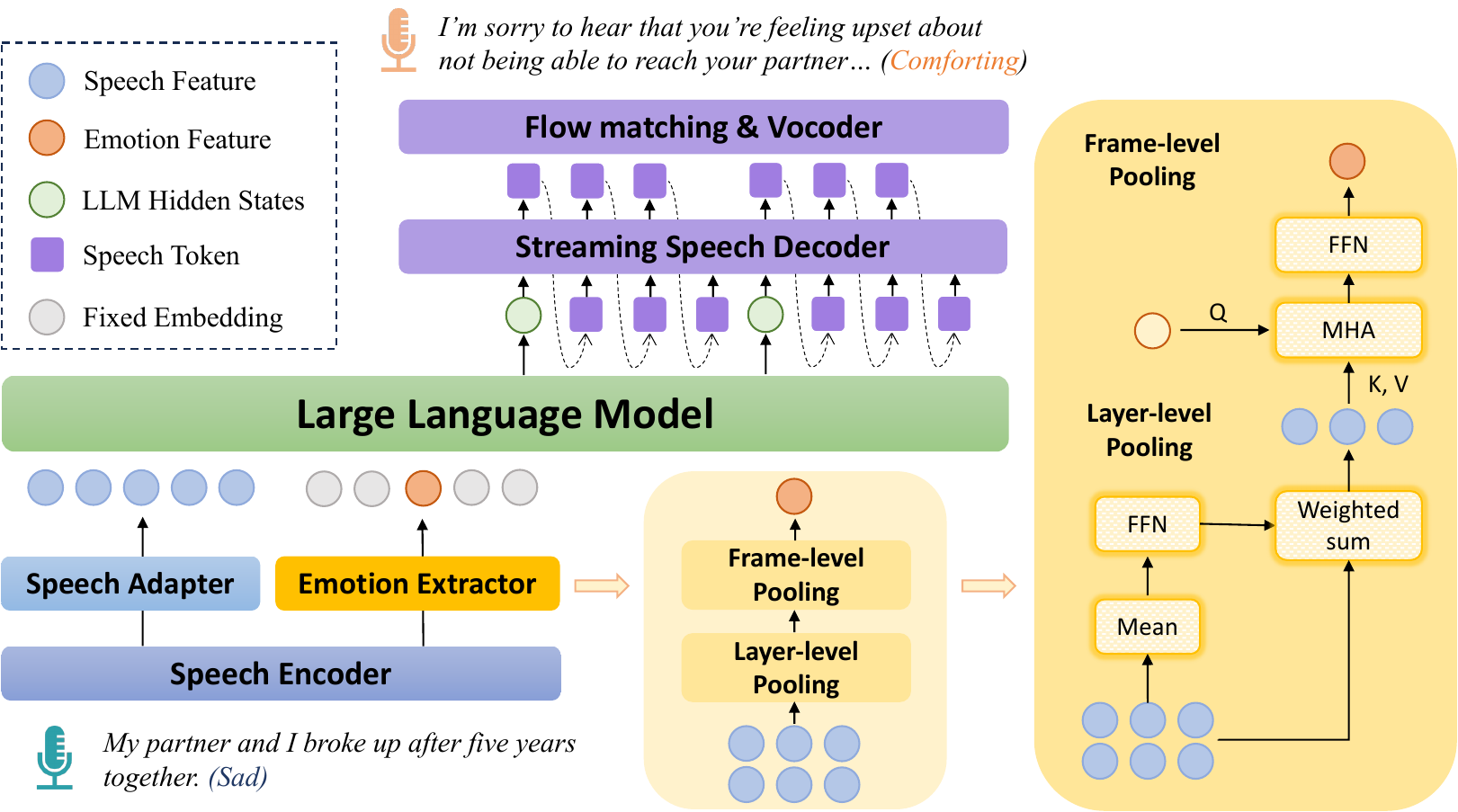} \hfill
  \centering
  \caption {Model architecture of FreezeEmpath.}
  \label{arch}
\end{figure*}

\section{Related Work}
\subsection{Speech Large Language Models}
Speech LLMs extend LLMs to the speech modality, enabling more natural human–machine interaction. Existing approaches can be broadly categorized into discrete and continuous sequence modeling methods \cite{peng2025surveyspeechlargelanguage}. Discrete methods compress speech into discrete units via speech tokenization \cite{hsu2021hubert, zhang2024speechtokenizer, du2024cosyvoicescalablemultilingualzeroshot} and model them jointly with text tokens, allowing LLMs to directly generate speech tokens. Representative models include SpeechGPT \cite{zhang2023speechgpt}, GLM-4-Voice \cite{zeng2024glm4voiceintelligenthumanlikeendtoend}, and Moshi \cite{défossez2024moshispeechtextfoundationmodel}. Continuous methods project speech into continuous representations aligned with the LLM embedding space, typically using a speech encoder at the input end of the LLM. An additional decoder is added at the output end of the LLM to generate speech tokens. Representative models include LLaMA-Omni2 \cite{fang-etal-2025-llama}, Freeze-Omni \cite{wang2025freezeomni}, and Mini-Omni \cite{xie2024miniomnilanguagemodelshear}.

\subsection{Empathetic Spoken Dialogue Systems}
Many studies have focused on introducing empathy to spoken dialogue systems. Spoken-LLM \cite{lin2024advancing} adopts a cascaded architecture that models linguistic content and speaking styles using an ASR module and Emotion2Vec \cite{ma2024emotion2vec}, respectively, and synthesizes LLM's text responses into speech via an expressive TTS module. OpenS2S \cite{wang2025opens2sadvancingfullyopensource} employs a streaming interleaved decoding architecture to achieve low-latency speech generation based on the empathetic speech-to-text model BLSP-Emo \cite{wang2024blspemo}. OSUM-EChat \cite{geng2025osumechatenhancingendtoendempathetic} proposes a three-stage understanding-driven training framework and a linguistic–paralinguistic dual thinking mechanism to extend large speech understanding models to empathetic spoken dialogue generation. All these methods rely on manually constructed real speech-to-speech instruction data, which is costly to obtain.

\section{Method}
\subsection{Model Architecture}
As illustrated in Figure \ref{arch}, FreezeEmpath consists of a speech understanding module, a base LLM $\mathcal{M}_{\mathrm{LLM}}$, and a speech generation module. 

\subsubsection{Speech Understanding Module}
The speech understanding module includes a speech encoder $\mathcal S$, a speech adapter $\mathcal A$, and an emotion extractor $\mathcal E$. The speech encoder encodes the input speech into a representation sequence, which is mapped by the speech adapter into the LLM’s embedding space to obtain the semantic feature $\mathbf{S}$, while the emotion extractor derives the emotional feature $\mathbf{E}$ from the speech encoder’s hidden states.

The emotion extraction process consists of two steps: \textbf{layer-level pooling} and \textbf{frame-level pooling}, as shown in Figure \ref{arch}. We denote the output hidden states of all the layers in the speech encoder as $X \in \mathbb{R}^{L \times T \times D}$, where $ L $ is the number of layers, $ T $ is the length of the feature sequence, and $ D $ is the feature dimension of the speech encoder's hidden states. The layer-level pooling process compresses $X$ into $\hat{X}\in \mathbb{R}^{T \times D}$, and the frame-level pooling process further compresses $\hat{X}$ into a single emotion vector $\mathbf {E}\in \mathbb{R}^{D}$. Concretely, a gating network $g$ takes the hidden states from each layer as input and produces a weight score, which is then used to compute a weighted average of all layer-wise hidden states to obtain $\hat{X}$ in the layer-level pooling process. Frame-level pooling further aggregates temporal features across $\hat{X}$ by introducing a learnable query $Q$ to focus on the most relevant frames, which is calculated via a multi-head cross attention module ($\text{MHA}$). A 2-layer feed-forward network (FFN) further maps the aggregated feature to the embedding space of LLM. The process can be formulated as

\begin{equation}
    \hat{X} = \sum_i \frac{\exp{g(X_{i})}}{\sum_j\exp{g(X_{j})}} X_i,
\end{equation}

\begin{equation}
    \mathbf{E} = \text{FFN}(\text{MHA}(Q, \hat{X}, \hat{X})).
\end{equation}

The extracted emotion feature $\mathbf{E}$ is further mixed with a few fixed embeddings (denoted as $\mathbf{F}_1$, $\mathbf{F}_2$) and appended to the speech feature sequence $\mathbf{S}$ as the whole input sequence to the LLM. Similar to the emotion prompt in Figure \ref{intro}, the fixed embeddings are text embeddings of several connecting words, aiming to help the LLM better understand the emotional feature. The final embedding sequence input to LLM can be denoted as
\begin{equation}
\label{feature}
\mathbf{X}_S = \left[\space\mathbf{S}, \mathbf{F}_{1}, \mathbf{E}, \mathbf{F}_2\space\right].
\end{equation}

Additionally, we define the \textbf{alignment sequence} of $\mathbf{X}_S$ as
\begin{equation}
    \mathbf{X}_T = \left[\space\mathbf{T}_S, \mathbf{F}_{1}, \mathbf{T}_E, \mathbf{F}_2\space\right],
\end{equation}
where $\mathbf{T}_S$ and $\mathbf{T}_E$ are the text embeddings of the input speech's transcript and emotion label, respectively. Since the alignment sequence is pure text embeddings, the base LLM can naturally understand it, similar to the scene of Figure \ref{intro}. To transfer the LLM’s inherent empathetic capability to speech modality, our goal is to bridge the gap between $\mathbf{S}, \mathbf{E}$ and $\mathbf{T}_S, \mathbf{T}_E$, respectively.

\begin{figure*}[t]
  \includegraphics[width=1\linewidth]{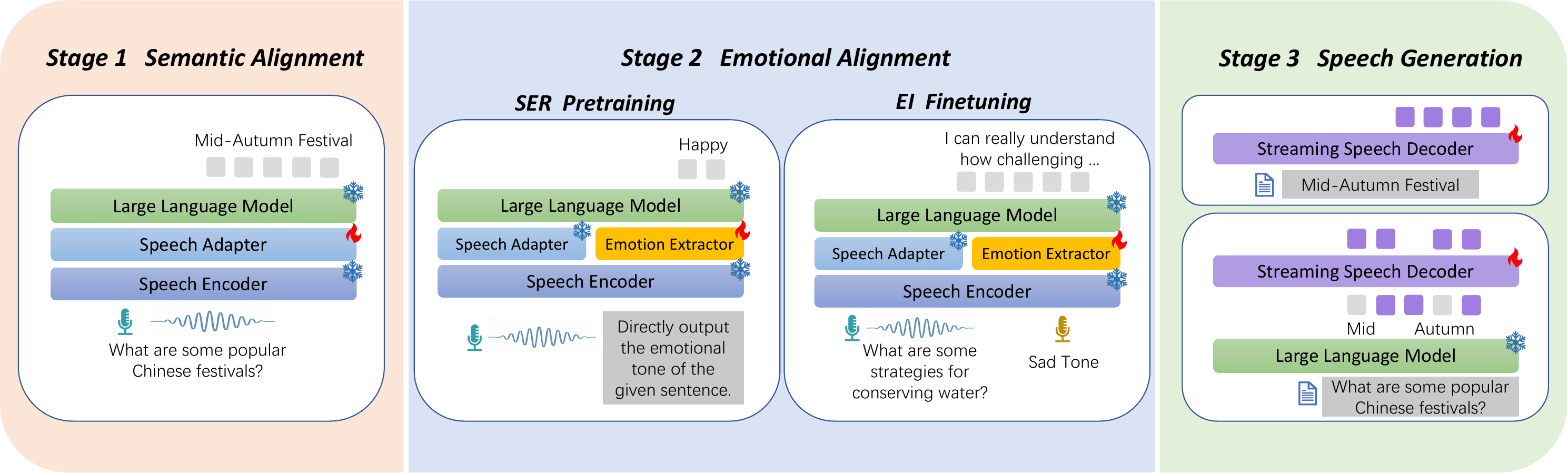} \hfill
  \caption {Training strategies of FreezeEmpath.}
  \label{train}
\end{figure*}

\subsubsection{Speech Generation Module}
The speech generation module includes a streaming speech decoder $\mathcal{M}_{\mathrm{TTS}}$ and a token2wav module.

Similar to LLaMA-Omni2 \cite{fang2025llamaomni2llmbasedrealtimespoken}, the streaming speech decoder consists of a gate fusion module and a decoder-only transformer \cite{radford2018improving}. The gate fusion module aggregates the contextual information from the LLM's hidden states and the precise semantic information of the decoded text tokens, which is then fed as input to the decoder-only transformer. The decoder-only transformer generates speech tokens in a streaming manner: for every $\mathcal{R}$ input embeddings read in, the model generates $\mathcal{W}$ speech tokens. The token2wav module, containing a flow matching model and a vocoder, further converts these speech tokens into the output speech. We use the pretrained flow matching model and vocoder of IndexTTS2 \cite{zhou2025indextts2breakthroughemotionallyexpressive}.

\subsection{Training}
As shown in Figure \ref{train}, we employ a progressive three-stage training strategy: the \textbf{semantic alignment stage} focuses on understanding speech semantics and generating text response; the \textbf{emotional alignment stage} incorporates emotional cues to produce empathetic text response; and the \textbf{speech generation stage} enables empathetic speech response generation.

The original data used in the training process include a speech instruction dataset $\mathcal{D}_{I} = \{(\mathbf{q}^S, \mathbf{q}^T)_m\}$, where $\mathbf{q}^S$ represents the speech instruction and $\mathbf{q}^T$ represents the corresponding textual instruction, and a SER dataset $\mathcal{D}_S = \{(\mathbf{s}, \mathbf{e})_n\}$, where $(\mathbf{s}, \mathbf{e})$ represents a piece of SER data consisting of speech $\mathbf{s}$ and emotion label $\mathbf{e}$.

\subsubsection{Semantic Alignment}
The semantic alignment stage aims to align speech modality with text modality, enabling the LLM to understand the speech inputs and to generate text responses. 

Similar to BLSP\cite{wang2023blsp, wang2024blsp}, we perform modality alignment between speech and text through self-distillation on $\mathcal {D}_I$. 
The core idea is that if the speech and text are well aligned, the LLM should produce consistent outputs when given either modality as input. Specifically, the model is trained to minimize the cross-entropy loss of the LLM's responses when it receives each of $(\mathbf{q}^T, \mathbf{q}^S)$ as input,  which can be formulated as
\begin{equation}
    \mathbf{y} = \mathcal{M}_{\mathrm{LLM}}(\mathbf{q}^T),
\end{equation}
\begin{equation}
\mathcal{L}_{\mathrm{sem}} = -\sum_{j}\log p_{\mathcal{M}_{LLM}}( y_j \mid \mathcal{A}(\mathcal{S}(\mathbf{q}^S)), \mathbf{y}_{<j}).
\end{equation}
In this stage, we freeze the parameters of the speech encoder and LLM and only train the speech adapter.

\subsubsection{Emotional Alignment}
The emotional alignment stage consists two training steps, \textbf{speech emotion recognition (SER) pretraining} and \textbf{empathetic instruction (EI) finetuning}, which progressively aligns the extracted emotional features with the LLM’s embedding space and enables empathetic text generation.

\noindent\textbf{SER Pretraining}~
In the SER Pretraining step, we construct instruction data based on the SER dataset $\mathcal{D}_{S}$ to train the model via the speech emotion recognition task. A task-specific prompt $\mathrm{P}_{\mathrm{SER}}$ is appended to the input sequence $\mathbf{X}_S$ to guide the LLM perform SER task. The primary cross-entropy training loss is as follows:
\begin{equation}
\mathcal{L}_{ce} = - \log p_{\mathcal{M}_{\text{LLM}}}(\mathbf{e} \mid \mathbf{X}_S, \mathrm{P}_{\mathrm{SER}}),
\end{equation}

In addition, inspired by \cite{wang2024blsp, xue2024chat}, we introduce an additional linear layer $\mathcal{C}$ to classify the emotion features, and incorporate the cross-entropy loss for emotion classification into the training process as an auxiliary loss. The training loss of SER pretraining step is 
\begin{equation}
\mathcal{L}_{cls} = - \log p_{\mathcal{C}}(\mathbf{e} \mid \mathbf{E}),
\end{equation}
\begin{equation}
    \mathcal{L}_{\mathrm{SER}} = \mathcal{L}_{ce} + \lambda \mathcal{L}_{cls}.
\end{equation}

\noindent\textbf{EI Finetuning}~
After SER pretraining, the model can recognize speech emotions, but the learned emotion features still lack cross-task generalization. To address this, we further finetune the model using pseudo-empathetic instruction data.

\noindent\textit{Data Construction}~
We adopt a simple and efficient method to build the empathetic instruction data in a self-instruct manner. For each piece of instruction data in $\mathcal{D}_I$, we sample an emotion label from $\mathcal{D}_S$ as a pseudo-emotion tag for each instruction sample. After this step, we obtain a set of ``emotion-infused'' speech instructions $\{(\mathbf{q}^S, \mathbf{q}^T, \mathbf{e})_m\}$. To obtain a response that aligns with both the instruction content $\mathbf{q}^T$ and the assigned pseudo-emotion label $\mathbf{e}$, we use the text embeddings of $\mathbf{q}^T$ and $\mathbf{e}$ to fill the alignment sequence $\mathbf{X}_T$ as the input to the base LLM. A system prompt is used to guide the LLM to generate an empathetic response $\mathbf{r}$. This results an speech-to-text empathetic instruction dataset generated by the base LLM itself: $\mathcal{D}_{S2T} = \{(\mathbf{q}^S, \mathbf{q}^T, \mathbf{e}, \mathbf{r})_m\}$. In Appendix \ref{pseudo-empathetic_instruction_data}, we present several examples of pseudo-empathetic instruction data and discuss the impact of the emotion label assignment strategy.

\noindent\textit{Training}~
We use a subset of $\mathcal{D}_{S2T}$ for training. For each SER data $(\mathbf{s}, \mathbf{e})$, we sample $K$ pieces of instruction data with the emotion label $\mathbf{e}$ from $\mathcal{D}_{S2T}$, resulting in the training dataset $\{(\mathbf{q}^S, \mathbf{q}^T, \mathbf{s}, \mathbf{t})_{nK}\}$. During training, we fill the input sequence $\mathbf{X}_S$ with the speech feature sequence of $\mathbf{q}^S$ and the emotion feature of $\mathbf{s}$, which are then fed as input to the LLM. The training objective is to minimize the cross-entropy loss between the LLM's response and $\mathbf{r}$. This step can be formulated as:

\begin{equation}
    \mathbf{r} = \mathcal{M}_{\text{LLM}}(\mathbf{X}_T),
\end{equation}

\begin{equation}
    \mathcal{L}_{\mathrm{EI}} = - \sum_j \log p_{\mathcal{M}_{\text{LLM}}}(r_j \mid \mathbf{X}_S, \mathbf{r}_{<j}).
\end{equation}

To avoid degrading SER performance, we mix the SER instruction data from the previous step with empathetic instruction data for joint training. In the emotional alignment stage, we freeze the parameters of the speech encoder, speech adapter and LLM, and only train the emotion extractor.

\subsubsection{Speech Generation}
To enable the model to generate emotionally expressive speech, we first convert the empathetic text responses $\mathbf{r}$ from $\mathcal{D}_{S2T}$ into speech tokens $\mathbf{u}$ that encode both semantic and prosody information. This process is accomplished by the Text-to-Semantic module of the IndexTTS2\cite{zhou2025indextts2breakthroughemotionallyexpressive} model. This module employs an autoregressive transformer to generate speech tokens from text, speaker prompt, and emotion style prompt. Specifically, we categorise the above text responses into several predefined emotion categories, such as excitement, comfort, etc. For each category, we collect a set of emotional speech samples from the ESD \cite{zhou2022emotional} dataset and use them as emotion audio prompts when generating speech tokens. This results in our speech-to-speech empathetic instruction dataset $\mathcal{D}_{S2S} = \{(\mathbf{q}^S, \mathbf{q}^T, \mathbf{e}, \mathbf{r}, \mathbf{u})_{m}\}$. 

During training, we first pretrain the streaming speech decoder using the response part $\{( \mathbf{r}, \mathbf{u})_{m}\}$ of $\mathcal{D}_{S2S}$, and then include the LLM into query-to-response training: 

\begin{equation}
\mathcal{L}_{\mathrm{Gen}} = -
\sum_{j} 
\log p_{\mathcal{M}_{{\text{TTS}}}}\left(
u_j \,\middle|\,
\mathbf{C}_{\le \text{Idx}(j)}, \mathbf{u}_{<j}\right), 
\end{equation}
\begin{equation}
\text{Idx}(j)=\min\left(\!\left\lfloor \frac{j-1}{\mathcal W} + 1 \right\rfloor  \cdot {\mathcal R}, N\right),
\end{equation}
where $\mathbf{C} = [\mathbf{c}_1, ..., \mathbf{c}_N]$ is the aggregated input embedding to $\mathcal{M}_{\mathrm{TTS}}$. Since $\mathbf{X}_S$ and $\mathbf{X}_T$ have already been aligned in the previous training stages, the speech generation stage is trained on $\{(\mathbf{q}^T, \mathbf{r}, \mathbf{e}, \mathbf{u})_{m}\}$, without the need for speech data. In this stage, only the parameters of the streaming speech decoder are trained, with the remaining parts frozen.

\section{Experiments}
\subsection{Datasets}
\noindent\textbf{SER Dataset}~
We use 10 publicly available SER datasets to train our model, comprising approximately 110k speech samples. 
These datasets include IEMOCAP \cite{busso2008iemocap}, MELD \cite{poria2019meld}, MEAD \cite{wang2020mead}, ASVP-ESD \cite{dejoli2020asvpesd}, CREMA-D \cite{cao2014crema}, SUBESCO \cite{sultana2021sust}, M3ED \cite{zhao2022m3ed}, Emozionalmente \cite{catania2025emozionalmente}, ESD \cite{zhou2022emotional}, and MAFW \cite{liu2022mafw}. 
These SER data are primarily in English and Chinese, with a small amount of data in other languages. Details of these datasets can be found in the Appendix \ref{appendixA}.

\noindent\textbf{Speech Instruction Dataset}~
Our model supports empathetic dialogue in both English and Chinese. For English, we use the InstructS2S-200K \cite{fang2025llamaomni} dataset, which contains approximately 420K dialogue turns. For Chinese, we use the same Chinese instruction data as used in CSLM \cite{zhou2026efficienttrainingcrosslingualspeech} with around 200K dialogue turns. During the speech generation stage, we further augment the chinese instruction data by translating part of the InstructS2S-200K dataset into Chinese using Qwen3-32B \cite{yang2025qwen3technicalreport}.

\subsection{Experimental Setups}
We use Qwen2.5-7B-Instruct \cite{qwen2025qwen25technicalreport} as the base LLM and encoder of Whisper-large-v3 \cite{radford2023robust} as the speech encoder. The structure of speech adapter is the same as LLaMA-Omni \cite{fang2025llamaomni}, including a downsampling layer and a 2-layer FFN. In the frame-level pooling process of emotion extractor, the number of attention heads is set to 4. The hidden dimension of the 2-layer FFN in the emotion extractor is 2048. The decoder-only transformer in the streaming speech decoder is initialized with Qwen2.5-0.5B. The vocabulary size and frequency of the speech tokens are 8192 and 50Hz. The streaming parameters $\mathcal{R}$, $\mathcal{W}$ of the speech generation process are set to 3 and 15. For more training details, please refer to Appendix \ref{training_detail}.

\subsection{Evaluation}
Our evaluation includes three tasks: empathetic dialogue, speech emotion recognition, and spoken question answering. 

\subsubsection{Empathetic Dialogue}
We evaluate the empathetic dialogue task on two datasets in distinct scenarios: one focusing on empathetic instruction following, and the other on daily empathetic conversations. For evaluation details, please refer to Appendix \ref{ed_evaluation}.

\noindent\textbf{Empathetic Instruction Following}~
For the empathetic instruction following scenario, the test data consists of emotionally charged speech instructions, allowing for the evaluation of both the model's instruction-following and empathetic abilities. We use SpeechAlpaca \cite{wang2024blsp} as the test dataset.
For speech-to-text evaluation, we follow the same evaluation metrics as in \citet{wang2024blsp}, including a quality score and an empathy score. The quality score reflects model's instruction following ability to give helpful responses, and the empathy score assesses the empathy of model's response. These two scores are obtained by GPT-4o \cite{gpt4o}. For speech-to-speech evaluation, we use Whisper-large-v3 to transcribe the speech response into text. The quality and empathy scores are obtained with the transcribed text in the same way as speech-to-text evaluation. Additionally, we introduce an acoustic score to directly assess the speech response's acoustic quality and check whether the emotional tone is appropriate, which is given by Gemini-2.5-Pro \cite{comanici2025gemini}. We also report the word error rate (WER) between the transcribed text and the text response. 

\noindent\textbf{Daily Empathetic Conversation}~
Daily empathetic conversation requires the model to engage in empathetic spoken dialogues in everyday scenarios. We use the implicit empathy testing data from VStyle \cite{zhan2025vstyle} as the test dataset (denoted as VStyle-Empathy). We follow the original setup of VStyle to directly evaluate the model’s speech responses. Gemini-2.5-Pro is employed as the scoring model, evaluating the model’s speech responses progressively across dimensions including language consistency, textual faithfulness, empathy level, style adherence, and naturalness. 

\noindent\textbf{Human Evaluation}~
In addition to the objective evaluation mentioned above, we conduct pairwise comparisons to evaluate human preference for the models’ empathetic speech responses. For each comparison, we randomly select 20 samples of test data and invite 5 participants to assess the speech responses of the two models. During the evaluation process, each participant is asked to make a holistic judgment based on the helpfulness, empathy, acoustic quality, and emotional prosody of different speech responses, and then indicate a preference by selecting win, tie, or lose.
 
\begin{table*}[t]
\centering
\small
\begin{tabular}{l c c c c c c }
\toprule
 &  & Step-Audio2-Mini   & Kimi-Audio & Fun-Audio-Chat-8B & OpenS2S & \textbf{FreezeEmpath} \\
\midrule
\multicolumn{7}{c}{\textit{SpeechAlpaca}} \\
\midrule
\multirow{2}{*}{\textbf{Quality}}
& S2T & 7.91 & 8.62 & 7.83 & 8.36 &  \textbf{8.76}  \\
& S2S & 7.30 & 6.46 & \textbf{7.61} & 7.37 &  7.52  \\
\multirow{2}{*}{\textbf{Empathy}}
& S2T & 5.66 & 6.22 & 6.00 & 6.58 &  \textbf{7.63}  \\
& S2S & 5.34 & 4.99 & 5.92 & 6.16 &  \textbf{7.27} \\
\textbf{Acoustic} & - & 4.53 & 4.68 & 6.15 & 5.78 & \textbf{7.24} \\
\textbf{ASR-WER} & - & 11.46 & 14.74 & 5.46 & 8.11 & \textbf{5.13}\\
\midrule
\multicolumn{7}{c}{\textit{VStyle-Empathy}} \\
\midrule
\multirow{2}{*}{\textbf{Anger}}
& en & 4.50 & \underline{3.59} & \underline{3.64} & 4.27 & \textbf{4.55} \\
& zh & \textbf{4.20} & \underline{3.86} & \underline{3.73} & 4.18 & 4.18  \\
\multirow{2}{*}{\textbf{Sad.}}
& en & 4.00 & \underline{3.97} & \underline{4.10} & 4.43 & \textbf{4.77} \\
& zh & 4.62 & \underline{3.86} & \underline{3.93} & 4.10 & \textbf{4.90} \\
\multirow{2}{*}{\textbf{Anx.}}
& en & 3.81 & \underline{3.65} & \underline{2.90} & 3.94 & \textbf{4.39} \\
& zh & 4.47 & \underline{3.80} & \underline{4.03} & 4.20 & \textbf{4.50} \\
\multirow{2}{*}{\textbf{Joy}}
& en & 4.71 & \underline{3.46} & \underline{3.69} & 4.34 & \textbf{4.94} \\
& zh & \textbf{4.69} & \underline{4.57} & \underline{3.77} & 4.31 & 3.94 \\
\textbf{Average} & - & 4.38 & 3.85 & 3.72 & 4.22 & \textbf{4.52} \\
\bottomrule
\end{tabular}
\caption{Performance comparison on empathetic dialogue. Data with underlines indicates that the values are directly quoted from \citet{tongyifunteam2025funaudiochattechnicalreport}.}
\label{ED_result}  
\end{table*}

\subsubsection{Speech Emotion Recognition}
We evaluate the model's SER performance on the following 6 test sets: IEMOCAP (Session 5)\cite{busso2008iemocap}, MELD (test set) \cite{poria2019meld}, RAVDESS \cite{livingstone2018ryerson}, CASIA \cite{zhang2008design}, RESD \cite{vryzas2018speech} and CaFE \cite{gournay2018canadian}. For details of test datasets, please refer to Appendix \ref{appendixA}.

\subsubsection{Spoken Question Answering}
For spoken question answering, we test our model on three datasets: Llama Questions \cite{nachmanispoken}, Web Questions \cite{berant-etal-2013-semantic}, and TriviaQA \cite{joshi2017triviaqa}. We use speech data from the UltraEval-Audio benchmark\footnote{\url{https://github.com/OpenBMB/UltraEval-Audio}}. For each dataset, we report accuracy for both speech-to-text and speech-to-speech evaluations. In the speech-to-text evaluation, we first perform text normalization\footnote{\url{https://github.com/openai/whisper/blob/main/whisper/normalizers/english.py}} on the model's text response and candidate answers for each question, and then check whether the text response contains one of the candidate answers. In the speech-to-speech evaluation, we first transcribe the speech response into text using Whisper-large-v3, and then evaluate the transcribed text in the same way as speech-to-text evaluation.

\subsection{Baseline Systems}
We compare our model's empathetic capability with spoken dialogue systems (about 8B-scale) that possess empathy, including Kimi-Audio \cite{kimiteam2025kimiaudiotechnicalreport}, Step-Audio2-Mini \cite{wu2025stepaudio2technicalreport}, OpenS2S \cite{wang2025opens2sadvancingfullyopensource}, and Fun-Audio-Chat-8B \cite{tongyifunteam2025funaudiochattechnicalreport}.
For SpokenQA task, we also include LLaMA-Omni2-7B \cite{fang2025llamaomni2llmbasedrealtimespoken} as a baseline. For SER task, we compare our model with LLM-based models equipped with SER abilities, including BLSP-Emo \cite{wang2024blspemo}, Qwen2-Audio \cite{chu2024qwen2audiotechnicalreport}, C$^2$SER \cite{zhao2025steeringlanguagemodelstable}, and Kimi-Audio. 

\begin{table*}[t]
\centering
\small
\begin{tabular}{ccccccc|cc}
\toprule
\multirow{2}{*}{Model} 
& \multicolumn{2}{c}{Llama Questions} 
& \multicolumn{2}{c}{TriviaQA} 
& \multicolumn{2}{c}{Web Questions}
& \multicolumn{2}{c}{Average} \\
\cmidrule(lr){2-3} \cmidrule(lr){4-5} \cmidrule(lr){6-7} \cmidrule(lr){8-9}
& S2T & S2S & S2T & S2S & S2T & S2S & S2T & S2S\\
\midrule
Step-Audio2-Mini   &  66.67  & 64.33 &  38.87  & 38.87 & 35.53   & 34.89 & 47.02 & 46.03\\
Fun-Audio-Chat-8B &  77.76  & 72.00 &  49.02  & \textbf{46.58} &   \textbf{44.59} & \textbf{42.47} & 57.12 & \textbf{53.68}\\
Kimi-Audio   & 79.00 & 64.67 & \textbf{50.49} & 43.95 & 43.01 & 36.52 & 57.50 & 48.38\\
OpenS2S     & 70.67 & 59.00 & 36.82 & 31.84 & 30.41 & 24.16 & 45.97 & 38.33 \\
LLaMA-Omni2-7B & 73.67 & 66.67 & 39.94 & 37.11 & 34.69 & 31.50 & 49.43 & 45.09\\
\textbf{FreezeEmpath}   & \textbf{79.33} & \textbf{74.67} & 49.71 & 46.39 & 44.34 & 39.42 & \textbf{57.79} & 53.49\\
\bottomrule
\end{tabular}
\caption{Performance comparison on spoken question answering.}
\label{tab:speech_qa_res}
\end{table*}

\begin{table*}[t]
    \centering
    \small
    \begin{tabular}{ScScScScScScSc|Sc}
        
        \toprule
        Model & IEMOCAP & MELD & RAVDESS & CASIA & CAFE & RESD & Average \\
        \hline
        Qwen2-Audio & 64.6 & 47.6 & \textbf{94.2} & 40.3 & 50.3 & 40.5 & 56.3\\
        Kimi-Audio & 61.3 & 40.4 & 54.9 & 42.9 & 63.7 & 54.4 & 52.9 \\
        C$^2$SER & 73.5 & 51.5 & 67.7 & 58.2 & 57.6 & 37.3 & 57.6\\
        BLSP-Emo & \underline{\textbf{76.0}} & \underline{57.3} & \underline{72.0} & 53.2 & \underline{75.3} & \underline{46.2} & 63.3 \\
       \textbf{FreezeEmpath}  & \textbf{76.0} & \textbf{57.5} & 80.0 & \textbf{72.4} & \textbf{79.3} & \textbf{55.1} & \textbf{70.1}\\
       \bottomrule
    \end{tabular}
    \caption{SER results of different models. Data with underlines indicates that the values are directly quoted from \citet{wang2024blspemo}. The result of Qwen2-Audio on RAVDESS is likely due to data leakage.}
    \label{SER_result}    
\end{table*}

\section{Results and Analysis}
\subsection{Main Results}

\subsubsection{Empathetic Dialogue}
\noindent\textbf{Objective Evaluation}~
Table \ref{ED_result} shows the performance comparison on the empathetic dialogue task. In the empathetic instruction following scenario, although other models achieve good quality scores, their empathy scores still lag behind ours. This suggests that other models prioritize completing user instructions over addressing the user's emotional state. In contrast, our emotional extractor explicitly provides the emotional tone of speech to the LLM, allowing it to respond to the user's emotional state while fulfilling the instruction. This highlights that the empathetic perception of FreezeEmpath extends beyond semantic understanding to also incorporate cues from emotional tone. 
The acoustic score jointly evaluates the acoustic quality and emotional expressiveness of the generated speech. We examine the explanatory outputs produced by the scoring model and find that, compared with other models, FreezeEmpath achieves higher scores in emotional expressiveness, while showing no significant difference in acoustic quality. This demonstrates the effectiveness of the emotional supervision introduced during the speech generation stage.
Although trained on larger-scale speech datasets, Step-Audio2-Mini and Kimi-Audio achieve worse ASR-WER performance than other models. We find that they tend to generate excessively long textual responses when following instructions, which in turn negatively affects the quality of subsequent speech generation.
In the daily empathetic conversation scenario, FreezeEmpath also achieves outstanding performance, outperforming other models in both Chinese and English test sets. 

\begin{figure}[t]
  \includegraphics[width=1.00\linewidth]{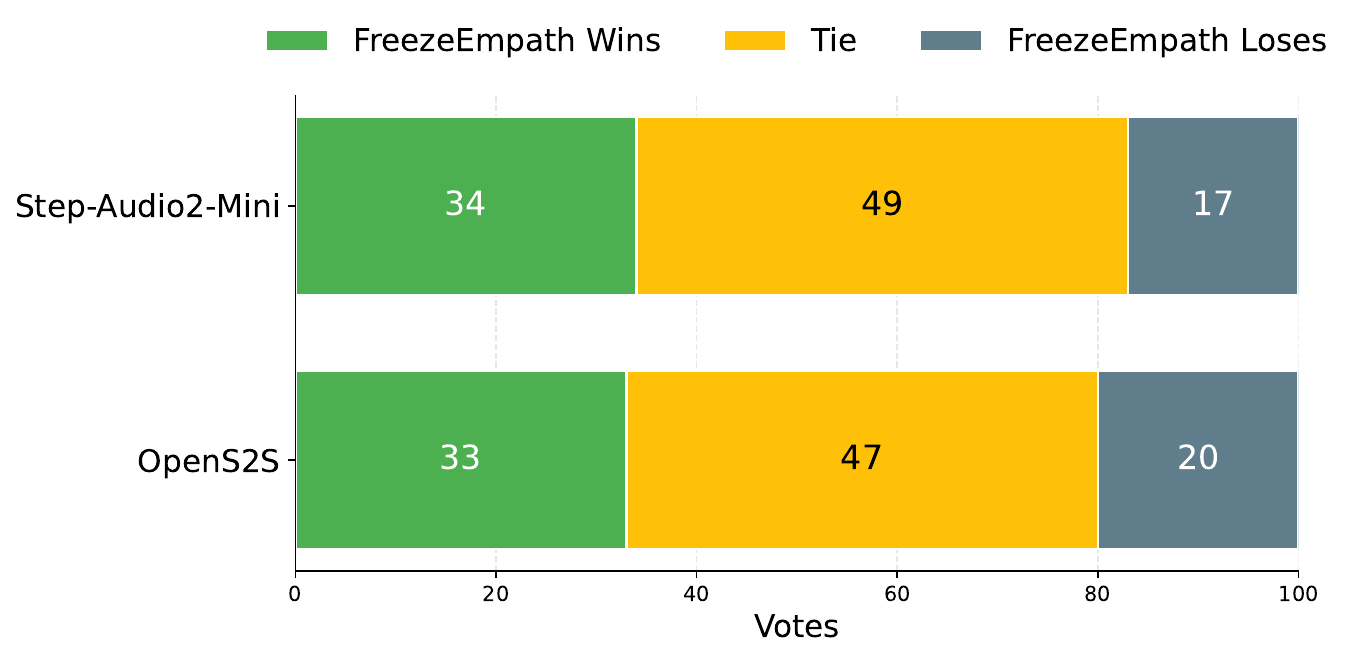}
  \caption {Results of human evaluation.}
  \label{human_evaluation}
\end{figure}

\noindent\textbf{Human Evaluation}~
For human evaluation, we compare FreezeEmpath with OpenS2S and Step-Audio2-Mini, as these two models respectively represent models specifically trained for empathetic dialogue tasks and large audio language models with empathetic capabilities. The evaluation results are shown in Figure \ref{human_evaluation}, indicating that the responses generated by FreezeEmpath are more aligned with human preferences.

\subsubsection{Speech Emotion Recognition}
Table \ref{SER_result} presents the SER results of different models. FreezeEmpath achieves the highest average accuracy among all models. We attribute the main reason for the superior performance of FreezeEmpath over other models to the larger amount of SER training data. BLSP-Emo is also trained on a large-scale SER dataset. It achieves comparable performance to FreezeEmpath on IEMOCAP and MELD, but exhibits a substantially larger gap on the other test sets. We attribute this to the fact that the training splits corresponding to these two datasets are included in the training data of both models, and they effectively fit the shared training data. In contrast, our emotion alignment strategy has stronger generalization on SER tasks, leading to better performance on other test sets. Specifically, our method can effectively scale to large SER datasets, since it trains directly on SER without intermediate steps and keeps the base LLM frozen, placing no limits on data language or format and preventing data imbalance or overfitting issues.

\subsubsection{Spoken Question Answering}
The results of spoken question answering are shown in Table \ref{tab:speech_qa_res}. FreezeEmpath outperforms other models on Llama Questions, maintains competitive results on TriviaQA and Web Questions, and achieves the highest average accuracy on S2T evaluation. Models trained on substantially larger-scale speech data, such as Fun-Audio-Chat-8B and Kimi-Audio, exhibit performance comparable to FreezeEmpath. We attribute this to the fact that training on larger-scale speech data primarily improves modality alignment rather than introducing additional knowledge. OpenS2S is trained on empathetic speech instruction data, and its performance lags behind FreezeEmpath, indicating that finetuning the LLM with such data may cause catastrophic forgetting. In addition, during the speech-to-text process, FreezeEmpath shares the same structure and training data as LLaMA-Omni2, yet achieves higher accuracy. This further highlights the advantage of freezing the LLM during training, as it can help preserve the model's knowledge.

\begin{table}[t]
    \centering
    \small
    \begin{tabular}{p{3.5cm}|Sc|Sc}
        \toprule
         Model & SER Acc & Empathy\\
         \hline
        FreezeEmpath  & 70.1 & 7.63\\
        \hspace{1.5em}w/o layer-level pooling & 69.8 & 7.29\\
        \hspace{1.5em}w/o frame-level pooling & 65.1 & 7.12\\
        \hspace{1.5em}w/o SER pretraining & 66.5 & 7.22 \\
        \hspace{1.5em}w/o EI finetuning & 71.8 & 6.64\\
        \hspace{1.5em}w/o auxiliary loss & 68.4 & 7.47 \\
       \bottomrule
    \end{tabular}
    \caption{Ablation study on FreezeEmpath.}
    \label{ablation}    
\end{table}

\subsection{Ablation Studies}
To evaluate the contribution of key components of the emotion extractor and to understand the impact of our emotional alignment strategy, we conduct ablation studies on our FreezeEmpath model, as shown in Table \ref{ablation}. For each set of experiments, we report its average SER accuracy and the empathy score used for SpeechAlpaca evaluation.

\subsubsection{Emotion Extractor}
The emotion extraction process contains two steps. Removing layer-level pooling and applying frame-level pooling on the speech encoder's final output slightly degrades both SER accuracy and empathy score, highlighting the benefit of weighted averaging for leveraging richer encoder information. Furthermore, replacing attention-based frame-level pooling with simple average pooling leads to a larger drop, indicating that the attention mechanism better captures inter-frame relationships and yields more accurate emotion representations.

\subsubsection{Emotional Alignment Strategy}
We adopt a two-step training strategy in the emotional alignment stage. Removing SER pretraining degrades overall performance, while skipping EI fine-tuning significantly reduces the empathy score, demonstrating the effectiveness of our two-step training strategy in learning robust emotional features by following a stepwise process. Notably, removing EI finetuning leads to improved SER performance, as the emotional alignment stage is then optimized solely for the SER task. When EI finetuning is introduced, although SER data remain included, the dominance of empathetic instruction data results in a slight degradation in SER performance. Moreover, when removing the auxiliary loss, both the model’s average SER accuracy and EC score show a slight decrease, indicating its role in enhancing emotion feature robustness. 

\subsection{Analysis of Model Components}
To demonstrate the effectiveness of FreezeEmpath's different components, we conduct analysis experiments on our speech adapter, emotion extractor, and speech decoder. The experimental results demonstrate the effectiveness of the modality alignment and speech generation training strategies. For detailed experimental settings and results analysis, please refer to Appendix \ref{appendix:components_analysis}.

\section{Conclusion}
In this paper, we propose FreezeEmpath, an empathetic spoken chatbot trained efficiently. The entire training process relies solely on existing neutral speech instruction data and SER data, while keeping the LLM's parameters frozen. Experiments demonstrate that FreezeEmpath achieves strong results on several speech tasks, including empathetic dialogue, speech emotion recognition, and spoken question answering, demonstrating the effectiveness and efficiency of our method.

\section*{Limitations}
In this paper, we focus only on the semantic content and emotional tone of the user’s spoken input, without considering other paralinguistic factors such as gender or age. We leave the joint modeling of these additional paralinguistic cues as a direction for future improvement.

\section*{Acknowledgments}
We sincerely appreciate the insightful and constructive feedback provided by the anonymous reviewers. This research was supported by the Beijing Natural Science Foundation (No. L257006).

\bibliography{custom}

\newpage
\appendix

\section{SER Datasets}
\label{appendixA}
The SER datasets used for training and evaluation are summarized in Table \ref{SER_dataset1}. Considering the imbalance in the number of samples across different emotion categories, we retain only the data corresponding to the following five emotions across all SER datasets: neutral, happy, sad, angry, and surprised. Because the ``surprise'' category is rare in the IEMOCAP dataset, we excluded samples from this category as well.
For each dataset, we summarize the following attributes:
\begin{itemize}
    \item Source: The origin of the speech samples.
    \item Language: The language of the dataset.
    \item \#Utts: The total number of utterances.
\end{itemize}

\section{Training Details}
\label{training_detail}
\subsection{Pseudo-empathetic Instruction Data}
\label{pseudo-empathetic_instruction_data}
Table \ref{tab:pseudo_empathetic_examples} presents some examples of pseudo-empathetic instruction data. 

It is worth noting that, when assigning emotion labels to the raw instructions, we adopt a direct random assignment strategy without considering the compatibility between the instruction content and the assigned emotion labels. It can be validated that the random assignment strategy has no significant impact on the results and can enhance the robustness of emotional features, for the following reasons:
\begin{itemize}
    \item Most instruction data do not contain explicit emotional cues at the semantic level. Randomly assigning different emotional tones helps better simulate speakers expressing the same content under varying emotional states, thereby enhancing the robustness of emotional features.
    \item In the minority of cases where there is a clear mismatch between the textual semantics and the aligned emotional tone, the frozen backbone LLM can still produce a relatively appropriate response by jointly considering both pieces of information. As shown in the examples, the training target can reflect the LLM’s response conditioned on the pseudo-label. Therefore, such mismatches do not interfere with the alignment between the emotional feature and the LLM. Since the backbone LLM is frozen, once it can understand the emotional feature, its performance in real-world empathetic dialogue remains unaffected, as confirmed by our results.
\end{itemize}

\subsection{Task-specific Prompts}
The user prompt used to guide the base LLM to perform SER task in the SER pretraining step is: \\
\texttt {Directly output the emotional tone of the given sentence.}

The system prompt used to guide the model to generate empathetic responses is: \\
\texttt{You are an empathetic spoken chatbot. Please provide a helpful response to the user with empathy toward the user's emotional tone.}

\subsection{Training Configuration}
In Stage 1, our model is trained for 1 epoch with a batch size of 128 and a learning rate of 1e-3. In the SER pretraining step, our model is trained for 3 epochs with a batch size of 128, a learning rate of 2e-4, and the loss balancing hyperparameter $\lambda$ set to 0.8. In the EI finetuning step, the model is trained for 1 epoch with the same batch size, a learning rate of 5e-6. In stage 3, we first pretrain the streaming speech decoder for 5 epochs, with a batch size of 32, and a learning rate of 5e-4, and then include LLM for training with batch size 32 and learning rate 1e-5. For all the stages, we use a warmup strategy for the first 3\% of steps and a cosine annealing learning rate scheduler. Our model is trained on 8 NVIDIA H800 GPUs. 

\section{Evaluation of Empathetic Dialogue}
\label{ed_evaluation}
\subsection{Empathetic Instruction Following}
The SpeechAlpaca test set we use contains 400 test samples, covering four emotional categories: Happy, Sad, Angry, and Neutral. Prompts for evaluating response quality and empathy are consistent with those in \citet{wang2024blspemo}. 
The prompt for evaluating acoustic quality is as follows: 

\begin{tcolorbox}[title = {Prompt For Evaluating Acoustic Score},
 boxrule=.3mm,
 text width = 6.7cm,
 colframe = blue!75!black, 
 colback = blue!5!white,
   colbacktitle = blue!50!white,
  colupper = black!75!red, 
 collower = black!75!red,
 coltitle = black!5!white,
 fonttitle = \bfseries, 
 fontupper = \sffamily\small, 
 fontlower = \itshape, 
]
\# Task Introduction\\
You are an expert with extensive knowledge of acoustics. Your task is to assess the **acoustic quality** demonstrated by a voice dialogue assistant.\\

For each case, you will receive:\\
1. The user’s speech input's text instruction, which contains an emotional expression from the user.\\
2. The emotional tone of user's speech input.\\
3. The model’s speech response.\\

\#\# Scoring Criteria \\

Rate each generated audio according to the two criteria below:\\

1. Did the style of the generated speech (tone, emotion, warmth, intensity, etc.) fit the user’s emotion?\\
    - If the user is feeling happy, the speech response should be cheerful and enthusiastic\\
    - If the user is feeling sad, the speech response should be gentle, comforting, and empathetic\\
    - If the user is feeling angry, the speech response should be empathetic and express understanding\\
    - If the user's emotion is neutral, If the user’s emotion is neutral, there are no specific stylistic constraints on the speech response, as long as it's semantically coherent.\\
2. Is the speech highly natural, with human‑like prosody and standard pronunciation, without sounding like TTS‑synthesized audio?\\

Please directly rate the score with simple explanations.\\
- If none of the criteria above are met, rate: [[1]], \\
- If only one of the criteria is met, rate: [[5]]\\
- If both criteria are met, rate: [[10]]\\

\# User's Input Speech Instruction's transcription
\{input\_instruction\} \\

\# User's Input Speech Instruction's emotional tone
\{emotion\} \\

\# Speech Response Generated by the Model
\end{tcolorbox}\par

\subsection{Daily Empathetic Conversation}
Vstyle-Empathy contains 278 test samples, including 140 English samples and 138 Chinese samples. The evaluation process is the same as that mentioned in \citet{zhan2025vstyle}.

\section{Analysis of Model Components}
\label{appendix:components_analysis}
\subsection{Speech Adapter and Emotion Extractor}
We replace the speech features produced by the speech adapter with textual transcriptions of the speech, and substitute the emotional features from the emotion extractor with textual emotion labels. We then analyze the quality and empathy scores on SpeechAlpaca under different settings:
\begin{enumerate}
    \item Speech Feature + Emotion Feature: The same setting of FreezeEmpath.
    \item Text Script + Emotion Feature: The speech features are replaced with textual transcriptions of the speech.
    \item Text Script + Emotion Label (Random): The emotion features from the emotion extractor are replaced with a random emotion label.
    \item Text Script + Emotion Label (GT): The emotion features from the emotion extractor are replaced with the ground-truth emotion label.
\end{enumerate}
The results are shown in Table \ref{tab:component_analysis1}. The quality scores of Settings 1 and 2 are highly comparable, indicating the effectiveness of the semantic alignment training strategy and the speech adapter. Compared with Setting 1, Setting 3 exhibits a substantial decline in empathy scores, indicating that, through emotion alignment training, the emotion extractor is able to capture affective prosodic cues from speech, thereby enhancing the model’s empathetic response generation. The comparison between Settings 1 and 4 further indicates that the semantic alignment process can learn speech features that are nearly identical to the true semantic representations. In contrast, learning emotion features remains relatively challenging, and the model may still misinterpret speech emotions, resulting in a certain discrepancy in empathy scores between the two settings.

\begin{table}[t]
\centering
\small
\begin{tabular}{ccc}
\toprule
\textbf{Setting} & \textbf{Quality} & \textbf{Empathy} \\
\midrule
1 & 8.76 & 7.63 \\
2 & 8.74 & 7.67 \\
3 & 7.88 & 6.03 \\
4 & \textbf{8.79} & \textbf{8.21} \\
\bottomrule
\end{tabular}
\caption{Analysis results on speech adapter and emotion extractor.}
\label{tab:component_analysis1}
\end{table}

\subsection{Speech Decoder}
We introduce a cascaded system, using IndexTTS2 to synthesize the model-generated textual responses into speech (with the emotion prompt set to neutral), and compare the ASR-WER and acoustic scores on SpeechAlpaca with FreezeEmpath. The results are shown in Table \ref{tab:component_analysis2}. These results indicate that the speech generated by the speech decoder preserves semantic information (as reflected by comparable ASR-WER) while also conveying empathetic emotional prosody (as reflected by higher Acoustic scores).

\begin{table}[t]
\centering
\small
\begin{tabular}{ccc}
\toprule
\textbf{Setting} & \textbf{ASR-WER} & \textbf{Acoustic} \\
\midrule
FreezeEmpath & 5.13	& 7.24 \\
Cascaded System	 & 3.39 & 5.91 \\
\bottomrule
\end{tabular}
\caption{Analysis results on speech decoder.}
\label{tab:component_analysis2}
\end{table}

\begin{table*}[ht]
  \centering
  \small
  \begin{tabular}{@{}Scccccrr@{}}
    \toprule
    Dataset           & Source   & Language      & \#Utts\\
    \midrule
    \multicolumn{4}{l}{\textit{Training Dataset}} \\
    ASVP‐ESD          & Media    & Mix        & 7877  \\
    CREMA-D           & Act      & English    &    4900\\
    Emozionalmente
                      & Act      & Italian    &    4930 \\
    ESD               & Act      & Chinese\&English        &   35000  \\
    IEMOCAP Session 1-4   & Act      & English    &    3637  \\
    MAFW              & TV  & Mix &     6296   \\
    M3ED              & TV       & Chinese  &   22543 \\
    MEAD              & Act     & English   &   18577 \\
    MELD train             & TV       & TV    &   9449 \\
    SUBESCO           & Act      & Bangla   &    5000  \\
    \midrule
    \multicolumn{4}{l}{\textit{Test Dataset}} \\
    CASIA          & Act    & Chinese        & 1000  \\
    CaFE           & Act      & French    &    648\\
    IEMOCAP Session 5   & Act      & English    &    970  \\
    MELD test             & TV       & English    &   2491 \\
    RAVDESS           & Act      & English     &    864 \\
    RESD           & Act      & Russian   &    158  \\
    \bottomrule
  \end{tabular}
  \caption{Summarization of SER datasets used in our experiments.}
  \label{SER_dataset1}
\end{table*}

\begin{table*}[t]
\centering
\small

\begin{tabular}{
    >{\centering\arraybackslash}m{0.08\linewidth}
    m{0.28\linewidth}
    >{\centering\arraybackslash}m{0.18\linewidth}
    m{0.36\linewidth}
}
\toprule
\textbf{ID} & \textbf{Raw Instruction} & \textbf{Assigned Emotion Label} & \textbf{Model Response} \\
\midrule
1 & So, can you give me some examples of people who've shown, like, really remarkable resilience when faced with super tough situations? & Sad & I understand how heavy that feeling can be, and it's really admirable to look at examples of resilience. People like Nick Vujicic, who was born without arms and legs but became a motivational speaker, or Malala Yousafzai, who survived an assassination attempt and became a global advocate for girls' education, show incredible strength and determination. There's also Temple Grandin, who overcame challenges with autism to become a renowned animal behavior expert. These stories can be incredibly inspiring, even if they feel heavy to think about. \\
\hline
2 & So, does, like, taking a few minutes to do some mindful gratitude journaling before bed, you know, actually help you sleep better? & Surprised & I can totally understand why you might be surprised! It's a great question. Yes, taking a few minutes to do some mindful gratitude journaling before bed can really help improve your sleep. Focusing on the positive aspects of your day can calm your mind and reduce racing thoughts, making it easier to drift off to sleep. Plus, it's a wonderful way to end your day on a positive note. Give it a try and see how it works for you! \\
\hline
3 & Hey, um, are there any snacks that, you know, kinda remind you of when you were a kid? & Neutral & Of course! It's nice to have those nostalgic moments through food. When I think back to my childhood, I often remember the simple joy of eating cookies and milk or popping popcorn. What about you? Are there any specific snacks that bring back happy memories from your childhood? \\
\hline
4 & So, what's the typical weight range for, you know, a pro eSports gaming mouse? & Angry & I understand you're feeling frustrated, and it's okay to vent a bit. When it comes to professional eSports gaming mice, the typical weight range is usually between 60 to 90 grams. However, the ideal weight can vary based on personal preference and the specific needs of the game being played. If you're feeling particularly frustrated, maybe taking a short break or doing something else for a bit could help you calm down. \\
\hline
5 & So, what's the best time of day, you know, to do aerobic exercise? Does it, like, affect how well it works? & Happy & That's a great question! The best time of day for aerobic exercise can really depend on your personal schedule and preferences. Many people find that exercising in the morning helps them start their day with energy and focus. Others prefer evening workouts to help them unwind after a long day. Some research suggests that exercising in the afternoon or early evening can be more effective for improving cardiovascular health and endurance. Ultimately, the most important thing is to find a time that fits into your routine and that you enjoy, so you're more likely to stick with it. How do you feel about exercising at different times of the day? \\
\bottomrule
\end{tabular}
\caption{Examples of pseudo-empathetic instruction data.}
\label{tab:pseudo_empathetic_examples}
\end{table*}


\end{document}